\documentclass[12pt,a4paper,english]{article}

\usepackage{censor}
\usepackage{ae,aecompl}
\usepackage[T1]{fontenc}
\usepackage[utf8]{inputenc}
\usepackage{geometry}
\usepackage{subcaption}
\usepackage{mathrsfs}
\usepackage{mdframed}
\usepackage{float}
\usepackage{tabularx}
\usepackage{babel}
\usepackage{float}
\usepackage{multirow}
\usepackage{lscape}
\usepackage{url}
\usepackage{mdframed}
\usepackage{afterpage}
\usepackage{indentfirst}
\usepackage{float}
\usepackage{longtable}
\usepackage{graphicx}
\usepackage{multirow}
\usepackage{pdflscape}
\usepackage{bm}
\usepackage{tabularx}
\usepackage{caption}
\usepackage{pifont}
\usepackage{setspace}
\usepackage{mathrsfs}
\usepackage{changepage}
\usepackage{tcolorbox}
\usepackage{dsfont}
\usepackage{algorithm}
\usepackage[noend]{algpseudocode}

\usepackage{natbib}
\usepackage{amsmath}
\usepackage{amssymb}
\usepackage{graphicx}
\usepackage{setspace}
\usepackage{changepage}

\makeatletter
\usepackage{babel}
\usepackage{bm}
\usepackage{setspace}
\makeatother

\newcommand{\oneVec}{{\bf 1}}

\newcommand{\rVec}{{\bf r}}
\newcommand{\RMat}{{\bf R}}
\newcommand{\xVec}{{\bf x}}
\newcommand{\xVecMC}{\xVec_{\text{MC}}}
\newcommand{\yVec}{{\bf y}}
\newcommand{\XVec}{{\bf X}}

\newcommand{\subs}[3]{#1_{#2},\ldots,\allowbreak #1_{#3}}

\newcommand{\XMC}{{\bf \mathcal{X}_{\text{MC}}}}
\newcommand{\XCand}{{\bf \mathcal{X}_{\text{cand}}}}
\newcommand{\XTrain}{{\bf \mathcal{X}}}
\newcommand{\yVecMC}{\yVec_{\text{MC}}}
\newcommand{\yVecHatMC}{\hat{\yVec}_{\text{MC}}}

\newcommand{\psiVec}{{\boldsymbol \psi}}

\DeclareMathOperator*{\argmin}{arg\,min}
\DeclareMathOperator*{\argmax}{arg\,max}

\begin{document}

\StopCensoring

\renewcommand{\thefootnote}{\fnsymbol{footnote}}

\begin{center}
{\large \textbf{Sequential Computer Experimental Design for Estimating an Extreme Probability or Quantile}}
\end{center}

\begin{center}
\textbf{\censor{HAO CHEN}}
\\
\emph{
\censor{Department of Statistics, University of British Columbia, Vancouver, BC, Canada V6T 1Z4} \\
{\tt \censor{hao.chen@stat.ubc.ca}}}
\end{center}

\begin{center}
\textbf{
\censor{WILLIAM J.\ WELCH}\footnotemark[1]} \\
\emph{
\censor{Department of Statistics, University of British Columbia, Vancouver, BC, Canada V6T 1Z4} \\ 
{\tt \censor{will@stat.ubc.ca}}}
\end{center}

\renewcommand{\thefootnote}{\fnsymbol{footnote}}

\footnotetext[1]{Corresponding author} 
\renewcommand{\thefootnote}{\arabic{footnote}}

\vspace*{2cm}
A computer code can simulate a system's propagation of variation
from random inputs to output measures of quality. Our aim here is to estimate a critical output tail probability or quantile 
without a large Monte Carlo experiment. Instead, we build a statistical surrogate for the input-output relationship with a modest number of evaluations and then sequentially add further runs, 
guided by a criterion to improve the estimate. We compare two criteria in the literature. Moreover, 
we investigate two practical questions: how to design the initial code runs and how to model the input distribution. 
Hence, we close the gap between the theory of sequential design 
and its application. \\


    
\newpage
\section{Introduction}
\label{sect:introduction}
Consider a deterministic computer model of an engineering system with $d$ inputs $ \xVec = (x_1, \ldots, \allowbreak x_d ) $ 
and output $  y = y(\xVec) $ measuring performance.
Although any one run of the computer code is deterministic,
to model manufacturing or end-use variation, the values of the inputs 
may be drawn from random variables $ \XVec = ( X_1, \ldots, \allowbreak X_d ) $.
Hence, we can define $Y = y(\XVec)$ as a random output.
This propagation of variation from $\XVec$ to $Y$ is our focus.
Specifically, without loss of generality, suppose $y$ is a smaller-the better output,
and $ y_f $ is a given critical value, where failure occurs say,
and we want to estimate the (small) failure probability $ p_f = \textrm{Pr}(y(\XVec) > y_f) $.
Quantile estimation is a similar inverse problem:
for given small probability $p_f$, find $y_f$ such that $ \textrm{Pr}(y(\XVec) > y_f)= p_f $. 
The focus of the paper, then, is to provide a good estimate of an extreme tail probability $p_f$ or an extreme quantile $y_f$.
We have in mind a computer model that is too expensive to run a simple Monte Carlo experiment.     

The motivating computer model in the paper is a model of a floor system. The inputs are $ d $ values of the modulus of elasticity (MOE) of the $ d $ floor joists and the output is the maximum deflection under a static load. If the maximum deflection exceeds a critical cut-off, the system will fail. The application will be described further in Section~\ref{sect:floor:model}. 


All methods considered are based on the Monte Carlo (MC) method but with implicit values of the output.
First, consider the straightforward MC problem, where it is feasible to evaluate the computer model many times 
and directly estimate $p_f$ or $y_f$. 
There is a large MC set 
$ \XMC = \{\xVecMC^{(1)}, \ldots,\xVecMC^{(N)}\} $ sampled from the distribution of $\XVec$. 
Plugging the $ \XMC$ points into the computer model, 
one can obtain the $ N $ outputs $ \yVecMC = \{y(\xVecMC^{(1)}), \ldots, y(\xVecMC^{(N)}) \}$. 
Then the empirical distribution of the output $ Y $ is given by 
\begin{equation} \label{eqn:mc}
\hat{F}(w) = \frac{1}{N}\sum\limits_{i = 1}^{N} \mathds{1} ( y(\xVecMC^{(i)}) < w ),
\end{equation} 
where $\mathds{1}(E) = 1$ if event $E$ is true and 0 otherwise. 
The failure probability or quantile are then the (approximate) solutions 
of $ p_f = 1 - \hat{F}(y_f)$. 
According to the Glivenko-Cantelli theorem \citep{Can, Gli}, 
the supremum of the difference between the empirical distribution and the true CDF converges to 0 almost surely. 

However, the computer code may be too computationally expensive to obtain a large MC sample. 
Consider a limited experimental budget, for example $  40 $ runs of the code. The empirical distribution of $ Y $ 
based directly on $ 40 $ runs would be inaccurate, especially in the tails.
The solution is to use a Gaussian process (GP) model \citep{sack1989, Willbook}, trained with a modest number of runs,
as a fast statistical proxy for the expensive computer model. The key ideas are as follows.
First, generate a discrete training set $ \XTrain = \{\xVec^{(1)}, \ldots,\xVec^{(n)}\} $,
where $n$ is relatively small, and obtain $ \yVec = (y(\xVec^{(1)}), \ldots, y(\xVec^{(n)}) )^T$ by running the computer model. 
The available data enable one to build a GP model. Second, obtain predicted values in a much larger MC set from the trained GP model: $ \yVecHatMC = \{\hat{y}(\xVecMC^{(1)}), \ldots, \hat{y}(\xVecMC^{(N)}) \}$. The failure probability $ p_f $ is estimated by $1 -  \hat{F}(y_f)$, 
where the true outputs $ \yVecMC $ in~(\ref{eqn:mc}) are replaced by $ \yVecHatMC $. 
The estimate of the quantile can be obtained in an analogous way.

There are several ways to choose the training set for fitting the GP. The first, which we call a fixed design strategy, is to use up all of the design budget $ n $, to train one statistical surrogate, which is employed to predict the outputs of the points in the MC set. The fixed strategy is simple, but the estimation accuracy may not be good enough, especially for estimating an extreme tail probability or quantile. The second way, a sequential design strategy, is to use part of the design budget at the beginning to train an initial GP and sequentially add points into the design space (one at a time), guided by a search criterion until the budget has been exhausted. Each time a new point is added, both the surrogate model and the probability/quantile estimate are updated. Compared with the fixed design strategy, the second strategy is a dynamic process that allows new information to be added ``on the fly'', and thus should intuitively provide more accurate estimation. A better estimation outcome using sequential methodology is well recognized; see \cite{jones1998efficient} and \cite{Ranjan} for examples. 

The core of a sequential method is the search criterion. \cite{jones1998efficient} proposed the expected improvement  (EI) criterion based on an improvement function for global optimization, and EI has been popular during the past 20 years. In general, there are different improvement functions for different statistical objectives. \cite{Ranjan} introduced an improvement function for contour estimation, i.e., search for $ \xVec = ( x_1, \ldots, x_d ) $ such that $ y(\xVec) =a $, where $ a $ is the target level. \cite{roy2014estimating} used two different criteria for estimating a quantile where $p_f = 0.2$: the first one is also based on EI and used nearly the same improvement function as \cite{Ranjan} for estimating the region defining a quantile. 
The second criterion of \cite{roy2014estimating}, the so-called hypothesis testing-based criterion, optimizes a ``discrepancy'' between the current prediction and the target quantile at any untried input set. The details will be reviewed in Section \ref{sect:seq}.  


Taking such a sequential design approach, the focus of this article is to provide recommendations for some practical questions arising when implementing the sequential strategy. The questions, along with summaries of some findings,
are as follows. 
\begin{itemize}
\item There are two search criteria for quantile and probability estimation: EI and the hypothesis testing-based criterion. Which yields a more accurate result, especially for an extreme probability or quantile? \cite{roy2014estimating} compared the performance of sequential designs and fixed designs and showed sequential designs produced more accurate quantile estimates. However, they did not consider probability estimation. None of their examples covered extreme quantile estimation. In this paper, we explicitly contrast the performances of EI and the hypothesis testing-based criterion for quantile and probability estimation. We conclude that the hypothesis testing-based criterion has a faster convergence to its target and hence is preferred.

\item How to specify the input distribution for $\XVec$? The distribution of $ Y = y(\XVec) $ 
is determined by the distribution of $ \XVec $ by propagating variation through the deterministic computer model. 
We explore in Section \ref{sect:input} several different ways of modelling the input distribution. 


\item How to generate the MC set? A simple random sample from the input distribution is straightforward, 
but it might be computationally inefficient even with GP prediction. 
In Section \ref{mcset}, we adopt a stratified sampling scheme to generate a smaller MC set 
such that an extreme probability or quantile is well estimated.

\item How to select the initial design for training the starting GP? 
In Section \ref{sect:pro} we show that over-sampling the tails of the $\XVec$ distribution may be preferable. 

\item What is a suitable stopping criterion for a sequential algorithm? It is critically important to know when to stop in practice when the ``true'' probability or quantile are unknown. Based on the hypothesis testing-based criterion, we generate diagnostic plots in Section \ref{set:diagg} to check if the sequence has converged or not. 

\item How to estimate the unknown parameters of a GP for these purposes? 
We use the Bayesian method of \cite{chen2017flexible} instead of maximum likelihood estimation (MLE). 
EI and the hypothesis testing-based criterion both require the predictive standard deviation of the response at an untried point. 
Bayesian methods are able to fully quantify parameter estimation uncertainty and are therefore preferred for training a GP model and predicting in sequential design.    
\end{itemize}
Thus, the major contributions are to close the gap between the theory of sequential design and its practical use by answering the above questions.   
The rest of the paper is organized as follows. In Section \ref{sect:floor:model}, we introduce the computer model used to illustrate the methodology. Gaussian processes are reviewed in Section \ref{sect:GP}. Section \ref{sect:seq} defines the sequential algorithm and the aforementioned search criteria in more detail. Section \ref{sect:toy} presents an empirical study based on failure of a column, and Section \ref{sect:wood} revisits the motivating floor system. Section \ref{set:diagg} proposes a diagnostic for stopping.
Finally, Section \ref{sect:remarks} makes some concluding remarks.    
     
\section{Computer Model of a Floor System}\label{sect:floor:model}
\begin{figure}[htbp!]
\centering
\includegraphics[height=2.1in, angle=0]{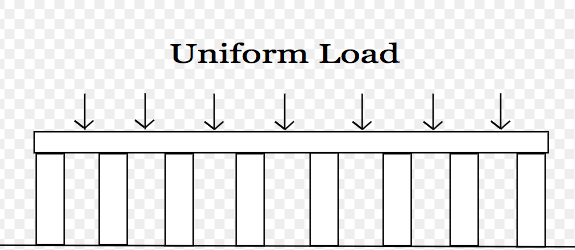}
\centering
\caption{Floor with $d = 8$ joists (supporting beams). 
The 8 joists act like springs, i.e., they deflect under a load.}
\label{fig:floor}
\end{figure}

\begin{figure}[htbp!]
\centering
\includegraphics[height=3.2in, angle=0]{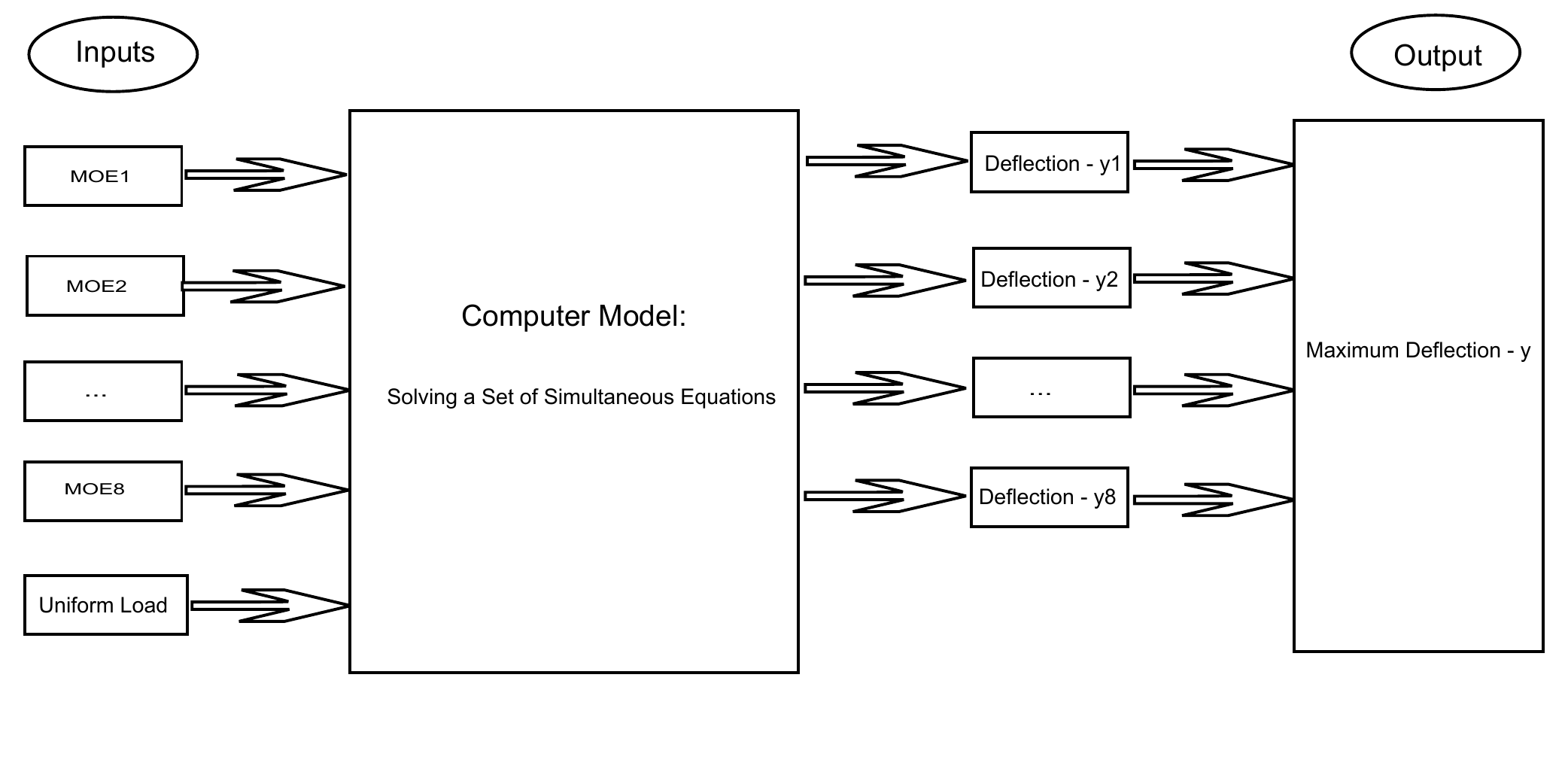}
\centering
\caption{Computer model of a floor system with $d = 8$ joists.}
\label{fig:floor:model}
\end{figure}

The application that motivates this research is a numerical model of a floor system \citep{wood}. 
The performance characteristic of interest is the floor's maximum deflection
under a static load. 
The $d$ inputs $\subs{x}{1}{d}$ to the model are the 
modulus of elasticity (MOE) values of $d$ supporting joists in units of psi.
For instance, Figure~\ref{fig:floor} shows a system with $d = 8$ joists.
The load could be a further input, 
but it will be kept constant in the analysis of Section~\ref{sect:wood} 
because we are more interested in the relationship between joist MOEs and the floor's maximum deflection. 
Given the inputs, the computer model outputs the $d$ deflections of the $d$ joists, 
and the maximum deflection is taken as the performance measure of interest, $y$ (units in);
see Figure~\ref{fig:floor:model}. 
The computer code will be treated as a black-box.

The code is deterministic,
but if we think of the inputs $\subs{x}{1}{d}$ as realizations of random variables $\subs{X}{1}{d}$,
then the output deflection $y$ is a realization of a random variable $Y$. 
Lumber is heterogenous relative to many other engineering materials, even within the same nominal grade,
and characterizing the induced uncertainty is important.
The reliability problem is to estimate either: 
\begin{itemize}
\item 
The failure probability $ p_f = \textrm{Pr}(Y > y_f)$, for a given maximum allowable deflection $ y_f $; or
\item  
The quantile $ y_f $ such that $ \textrm{Pr}(Y > y_f)= p_f $, where $ p_f $ is a given small probability. 
\end{itemize}

\section{Gaussian Processes and Parameter Estimation}\label{sect:GP}

\subsection{Gaussian Process Model} \label{sect:GP:GP}
We now briefly review the GP model. 
For the analysis of computer experiments \cite{sack1989} 
treated a deterministic output function $y(\xVec)$ as a realization of a GP,
\begin{equation}\label{GP}
\mu + Z(\xVec).  
\end{equation}
Here, $ \mu $ is the mean 
and $Z(\xVec)$ is a stationary GP with zero mean and unknown variance $ \sigma^2 $.
The mean could also be a regression function in the inputs $\xVec$,
but \cite{chen2016analysis} showed that a constant mean model is often sufficient.
Given two input vectors $ \xVec $ and $ \xVec'$, 
a power-exponential structure is assumed for the correlation between $Z(\xVec)$ and $Z(\xVec')$: 
\begin{equation} \label{eqn:cor:fn}
R(\xVec, \xVec') = \exp\left(-\sum\limits_{j = 1}^{d}\theta_j| x_j - x'_j|^{p_j}\right),  
\end{equation} 
where $\theta_j > 0 $ and $1 \le p_j \le 2$ for all $j$.

Conditional on the training data $\yVec$,
the correlation parameters $\psiVec = (\theta_1, \ldots, \theta_d, p_1, \ldots, p_d)$, $\mu$ and $\sigma^2$,
the predictive distribution for $y(\xVec^{*})$ at a new point $\xVec^{*}$ is Gaussian, 
i.e.,
\begin{equation}
\label{eqn:normal}
N \left(m(\xVec^{*}), v(\xVec^{*})\right),
\end{equation}
with predictive mean 
\begin{equation} \label{eqn:mean}
m (\xVec^{*}) = \mu + \rVec^T(\xVec^{*}) \RMat^{-1} (\yVec- \oneVec \mu),
\end{equation}
and predictive variance
\begin{equation}\label{eqn:var}
v (\xVec^{*}) = \sigma^2
\left( 1-\rVec^T(\xVec^{*})\RMat^{-1}\rVec(\xVec^{*})\right).
\end{equation}
Here, $ \rVec(\xVec^{*})$ is an $n \times 1$ vector obtained from (\ref{eqn:cor:fn}) 
with element $ i $ given by $ R(\xVec^{*}, \xVec^{(i)}) $ for $ i = 1,\ldots, n $,
the $ n \times n $ matrix $ \RMat $ has element $ i,j $ given by $ R(\xVec^{(i)}, \xVec^{(j)}) $ 
for $1\le i, j \le n $,
and $\oneVec$ is an $n \times 1$ vector of 1's.

\subsection{Parameter Estimation}\label{sect:GP:par}

The quantification of the total uncertainty of prediction is critical for the search criteria in Section \ref{sect:seq}. Hence, a Bayesian method accounting for parameter estimation uncertainty \citep{chen2017flexible} is used and briefly described here.

At any stage the computer model has been evaluated at $n$ input configurations
$\xVec^{(i)} $ for $ i = 1,\ldots, n $.
With a vague uniform prior on $\mu$ and an inverse-gamma prior on $\sigma^2$ 
with both the shape and scale parameters tending to zero,
the predictive distribution given all correlation parameters $\psiVec $ becomes
\begin{equation}\label{{eqn:pred:t}}
t_{n-1} \left(\hat{m}_{\psiVec} (\xVec^{*}), \hat{v}_{\psiVec}(\xVec^{*})\right),
\end{equation}
i.e., a shifted and scaled $t$ distribution with $n - 1$ degrees of freedom \citep{handcock}.
The predictive mean in (\ref{eqn:mean}) becomes 
\begin{equation} \label{eqn:mean2}
\hat{m}_{\psiVec} (\xVec^{*}) = \hat{\mu}_{\psiVec} + \rVec^T(\xVec^{*}) \RMat^{-1} (\yVec- \oneVec \hat{\mu}_{\psiVec}),
\end{equation}
where 
\begin{equation*}
\hat{\mu}_{\psiVec} = \frac{\oneVec^T \RMat^{-1}\rVec(\xVec^{*})}
{\rVec^T(\xVec^{*}) \RMat^{-1} \rVec(\xVec^{*})},
\end{equation*} 
and the predictive variance in (\ref{eqn:var}) becomes
\begin{equation} \label{eqn:var2}
\hat{v}_{\psiVec} (\xVec^{*}) = \widehat{\sigma^2}_{\psiVec} \left( 1-\rVec^T(\xVec^{*})\RMat^{-1}\rVec(\xVec^{*}) 
+ \frac{\left(1- \oneVec ^T\RMat^{-1}\rVec(\xVec^{*})\right)^2}{\oneVec^T \RMat^{-1}  \oneVec} \right),
\end{equation}
where
\begin{equation*}
\widehat{\sigma^2}_{\psiVec} = \frac{(\yVec - \oneVec \hat{\mu}_{\psiVec})^T \RMat^{-1} (\yVec - \oneVec\hat{\mu}_{\psiVec} )}{n - 1}.
\end{equation*}

Next, the correlation parameters $\psiVec$ are handled by Markov chain Monte Carlo (MCMC). 
The MCMC algorithm takes independent priors on the smoothness parameters $p_j$ 
as well as the $\theta_j$ as described by \cite{chen2017flexible} 
in their fully Bayesian implementation of the power-exponential structure.
Let $\psiVec^{(i)}$ for $i = 1,\ldots, M$ be the MCMC sample of size $M$ of the correlation parameters.
For each $\psiVec^{(i)}$, 
the conditional predictive mean and variance 
in (\ref{eqn:mean2}) and (\ref{eqn:var2}), respectively, are computed.
The overall predictive mean is the average of these means, 
\begin{equation}\label{eqn:mean:mcmc}
\hat{m} (\xVec^{*}) = \frac{1}{M}\sum\limits_{i=1}^{M} \hat{m}_{\psiVec^{(i)}}(\xVec^{*}), 
\end{equation}
and the overall predictive variance is obtained through the law of total variance,  
\begin{equation}\label{eqn:var:mcmc}
\hat{v} (\xVec^{*})= \frac{1}{M}\sum\limits_{i=1}^{M} \widehat{v}_{\psiVec^{(i)}}(\xVec^{*})
+ \frac{1}{M-1}\sum\limits_{i=1}^{M}\left(\hat{m}_{\psiVec^{(i)}}(\xVec^{*})-\hat{m}(\xVec^{*}) \right)^{2}. 
\end{equation} 
These Bayesian methods are used to compute the predictive mean and variance
for all the sequential-search criteria in Section~\ref{sect:seq}.


\section{Sequential Experimental Design}
\label{sect:seq}
Here we describe the sequential design strategies for estimation of a tail probability or quantile.
The heart of a sequential algorithm is the criterion for adding new evaluations to the search,
and we contrast existing criteria. 

\subsection{Sequential Algorithms}\label{sect:seq:alg}
The basic idea is to apply MC using the GP predictions of $y(\xVecMC^{(i)})$ for all points
in the MC set $\XMC$ rather than evaluating the computer model.
Given initial training data of $n_0$ evaluations, $\yVec = y(\xVec^{(1)}),\ldots, y(\xVec^{(n_0)})$,
and hence a trained GP, 
Algorithms~\ref{alg:p} and \ref{alg:q} compute tail probability and quantile estimates, respectively.
They add $ n_+ $ points sequentially according to the search criteria 
in Sections~\ref{sect:EI} and~\ref{sect:distance}.

\begin{algorithm}
\caption{Return an estimate of $ p_f = \textrm{Pr}(Y > y_f)$ for a given $y_f$}
\label{alg:p}
\begin{algorithmic}[1]
\Function{ProbabilityEstimate}{$n_0, n_+, \XTrain, \yVec, y_f$}
\State $n = n_0$        
\For{$i = 1$ to $n_+$}
\State Use the current training data, $\XTrain$ and $ \yVec $, to fit the GP   
\State Compute predictions $\hat{y}(\xVecMC^{(1)}), \ldots, \hat{y}(\xVecMC^{(N)})$ for the MC set 
\State $\hat{p}_f = (1 / N) \sum_{i = 1}^{N} \mathds{1} (\hat{y}(\xVecMC^{(i)}) > y_f)$
\If {$i < n_+$}   
\State Choose $\xVec^{(n+1)}$ from $\XCand$ based on a search criterion 
\State Append  $\xVec^{(n+1)}$ to $\XTrain$
\State Evaluate $y(\xVec^{(n+1)})$ and append it to $\yVec$
\State $n$ is replaced by $n + 1$
\EndIf 
\EndFor
\State \Return $\hat{p}_f$
\EndFunction
\end{algorithmic}
\end{algorithm}

\begin{algorithm}
\caption{Return an estimate of $y_f$, where $\textrm{Pr}(Y > y_f) = p_f$ for a given $p_f$}
\label{alg:q}
\begin{algorithmic}[1]
\Function{QuantileEstimate}{$n_0, n_+, \XTrain, \yVec, p_f$}
\State $n = n_0$        
\For{$i = 1$ to $n_+$}
\State  Use the current training data, $\XTrain$ and $ \yVec $, to fit the GP  
\State  Compute predictions $\hat{y}(\xVecMC^{(1)}), \ldots, \hat{y}(\xVecMC^{(N)})$ for the MC set  
\State  Compute $\hat{y}_f$ such that 
$p_f \simeq (1 / N) \sum_{i = 1}^{N} \mathds{1} (\hat{y}(\xVecMC^{(i)}) > \hat{y}_f)$ 
\If {$i < n_+$}   
\State  Choose $\xVec^{(n+1)}$ from $\XCand$ based on a search criterion
\State Append  $\xVec^{(n+1)}$ to $\XTrain$
\State Evaluate $y(\xVec^{(n+1)})$ and append it to $\yVec$
\State $n$ is replaced by $n + 1$
\EndIf
\EndFor
\State \Return $\hat{y}_f$
\EndFunction
\end{algorithmic}
\end{algorithm}
Note that in Section \ref{sect:wood}, the probability estimate in step 6 of Algorithm \ref{alg:p} will be replaced by a stratified random sampling estimate. Step 6 in Algorithm \ref{alg:q} for quantile estimation will be replaced in an analogous way. Both algorithms are based on the method of \cite{Ranjan} for contour estimation up to the search criterion.  
Mapping out a given contour where $y(\xVec) = y_f$ is equivalent to separating the MC points into 
those where $y(\xVecMC^{(i)}) > y_f$ versus those where $y(\xVecMC^{(i)}) \le y_f$.
In Algorithm~\ref{alg:p} it is straightforward to estimate $p_f$ \citep{BicMahEld2009}
with $\hat{y} (\xVecMC^{(i)})$ equal to $ \hat{m}(\xVecMC^{(i)})$ in (\ref{eqn:mean:mcmc}).
For quantile estimation, Algorithm~\ref{alg:q} estimates the unknown quantile 
at each iteration and then chooses the next point as if the estimated quantile 
is the known contour of interest \citep{roy2014estimating}

\subsection{Expected Improvement Criterion} \label{sect:EI}


For mapping out a contour where $y(\xVec)$ equals the constant $y_f$, 
the improvement function proposed by \cite{Ranjan} can be written as
\begin{equation} \label{eqn:I}
I(\xVec)= \begin{cases}
\alpha^2 v_{\psiVec,\sigma^2}(\xVec)-(y(\xVec) - y_f)^2  
& \mbox{if $ | y(\xVec) - y_f | < \alpha \sqrt{v_{\psiVec,\sigma^2} (\xVec)} $} \\
0  & \mbox{otherwise},      
\end{cases}
\end{equation}
where $ v_{\psiVec,\sigma^2} (\xVec)  $ is given by
\begin{equation*} 
v_{\psiVec, \sigma^2} (\xVec^{*}) = \sigma^2 \left( 1-\rVec^T(\xVec^{*})\RMat^{-1}\rVec(\xVec^{*}) 
+ \frac{\left(1- \oneVec ^T\RMat^{-1}\rVec(\xVec^{*})\right)^2}{\oneVec^T \RMat^{-1}  \oneVec} \right). 
\end{equation*} 
For the derivation of EI the predictive distribution $ y(\xVec) $ is taken to be normal, with known predictive variance. Hence, the two subscripts $\psiVec$ and $\sigma^2$ emphasize the predictive variance is conditional on these two GP parameters. The constant $\alpha$ determines the level of confidence. Thus, a large improvement would result from a new evaluation at $\xVec$ 
where the uncertainty measured by $v_{\psiVec,\sigma^2}(\xVec)$ is currently large and $y(\xVec)$ 
turns out to be close to the target $y_f$. Taking the expectation of $I(\xVec)$ with respect to the predictive distribution of $y(\xVec)$
gives EI,
\begin{equation}\label{eqn:EI}
\begin{array}{lcl} 
E (I(\xVec)) 
&=& \left(\alpha^2 v_{\psiVec,\sigma^2}(\xVec) - (\hat{m}_{\psiVec}(\xVec) - y_f)^2 \right)(\Phi(u_2)-\Phi(u_1)) \\
&+& 
v_{\psiVec,\sigma^2}(\xVec)\left( ( u_2 \phi(u_2) - u_1 \phi(u_1) )- (\Phi(u_2)-\Phi(u_1)) \right) \\
&+& 
2 (\hat{m}_{\psiVec}(\xVec) - y_f) \sqrt{v_{\psiVec,\sigma^2}(\xVec)} (\phi(u_2)-\phi(u_1)), 
\end{array}
\end{equation}
where $\hat{m}_{\psiVec}$ is the predictive mean in (\ref{eqn:mean2}),
$u_1 = (y_f - \hat{m}_{\psiVec}(\xVec)) / \sqrt{v_{\psiVec,\sigma^2}(\xVec)} - \alpha  $, 
$u_2 = (y_f - \hat{m}_{\psiVec}(\xVec)) / \sqrt{v_{\psiVec,\sigma^2}(\xVec)} + \alpha$, 
and $ \phi(\cdot) $ and $\Phi(\cdot)$ are the PDF and CDF of the standard normal distribution, respectively. 
In practice, the conditional mean $\hat{m}_{\psiVec}(\xVec)$ and variance $v_{\psiVec,\sigma^2}(\xVec)$
have to be replaced by their unconditional estimates in (\ref{eqn:mean:mcmc}) and (\ref{eqn:var:mcmc}), respectively.
The sequential-design criterion for selecting the next point, $\xVec^{*}$, based on EI is 
\begin{equation} \label{eqn:max:EI}
 \xVec^{*} =\argmax_{ \xVec \in \XCand} E(I( \xVec)),  
\end{equation}
where $ \XCand $ can be either a discrete or continuous candidate set.



\subsection{Discrepancy Criterion}\label{sect:distance}
\cite{roy2014estimating} defined the ``discrepancy'' between $y(\xVec)$ and $ y_f $ 
at untried input vector $\xVec$ to be 
\begin{equation*}
D(\xVec)= \begin{cases}
 \frac{(y(\xVec) - y_f)^2 + \epsilon}{v_{\psiVec,\sigma^2}} & \mbox{if $v_{\psiVec,\sigma^2} > 0$} \\
\infty & \mbox{otherwise},   
\end{cases}
\end{equation*}
where $ \epsilon > 0 $. If $ \epsilon = 0 $, the expression looks like an $F$-statistic. 
\cite{roy2014estimating} proposed choosing the next code evaluation at 
\begin{equation} \label{tg}
 \xVec^{*} = \argmin_{ \xVec \in \XCand} E(D(\xVec)),
\end{equation}
where the expectation is respect to the predictive distribution of $y(\xVec)$.
The expectation can be written as 
\begin{equation*}
E(D(\xVec))= \begin{cases}
\left( \frac{\hat{m}_{\psiVec}(\xVec) - y_f}{\sqrt{v_{\psiVec,\sigma^2}}} \right)^2 + \frac{\epsilon}{v_{\psiVec,\sigma^2}} + 1, & \mbox{if $v_{\psiVec,\sigma^2} \neq 0$} \\
\infty & \mbox{otherwise},   
\end{cases}
\end{equation*}
and, as we work with $ \epsilon = 0 $, an equivalent optimization is  
\begin{equation} \label{tg}
 \xVec^{*} = \argmin_{ \xVec \in \XCand} \left( \frac{|\hat{m}_{\psiVec}(\xVec)-y_f|}{\sqrt{v_{\psiVec,\sigma^2}( \xVec)}} \right).
\end{equation}    
where $ v_{\psiVec,\sigma^2}( \xVec) > 0 $ if the search omits points already in the training set. 
Another equivalent formulation is  
\begin{equation} \label{pro}
\xVec^{*} = \argmax_{ \xVec \in \XCand} \textrm{Pr} \left(Z< -\frac{|\hat{m}_{\psiVec}( \xVec)-y_f|}{\sqrt{v_{\psiVec,\sigma^2}( \xVec)}}\right),
\end{equation}
where $ Z \sim N(0, 1)$.
We see that the discrepancy criterion is related to EI in (\ref{eqn:EI}),
but it does not require a confidence-level $\alpha$.
Like the maximization of EI in (\ref{eqn:max:EI}), 
maximizing (\ref{pro}) balances choosing a point with predictive mean close to $ y_f $
(local search) and choosing a point with large predictive variance (global search).

\section{Example: Short Column Function}
\label{sect:toy}

To illustrate the sequential criteria and to start to address some questions about
their practical implementation, we take a model of a short column with uncertain material properties under an uncertain load. It was used by \cite{reliability} to study the trade-off between structural reliability and cost; we consider only the reliability component of the function. 
For a given input $ \xVec= (x_m, x_p, x_z) $ it is 
\begin{equation} \label{d3_e}
y(\xVec)=1-\frac{4 x_m}{b h^2 x_z}-\frac{x_p^2}{b^2 h^2 x_z^2}, 
\end{equation}
where the output $ y(\xVec) $ is the limit state, i.e., the difference between resistance and load, 
and $ y(\xVec) < 0$ means the system fails.
The three inputs ($ x_m, x_p, x_z $) are sampled from their respective independent distributions \citep{reliability}: 
\begin{itemize}
\item $X_m \sim N(\text{mean}=2000, \text{sd}=400) $ is the bending moment;   
\item $X_p \sim N(\text{mean}=500, \text{sd}=100) $ is the axial force; 
\item $X_z \sim \text{Lognormal} (\text{mean}=5, \text{sd}=0.5) $ is the yield stress.  
\end{itemize}
The parameters $ b $ and $ h $ are the width and depth of the cross-section, respectively,
both in mm.
Because the function is trivial to compute in this illustration, 
it is possible to simulate $ 10 $ million points 
from the input distributions and establish that taking $ b=3$ and $h=10$ gives $ \textrm{Pr}(y(\XVec) < 0)=0.0025 $ with negligible binomial standard error. 
Therefore, the true probability of system failure is $ 0.0025$, 
which we aim to estimate using sequential methodology. 

Note that the probability of interest in the short column function is actually a lower-tail probability. Hence step 6 of Algorithm \ref{alg:p} will change to 
$ \hat{p}_f = (1 / N) \sum_{i = 1}^{N} \mathds{1} (\hat{y}(\xVecMC^{(i)}) < 0) $. The quantile estimation in Algorithm \ref{alg:q} will change in a similar way. 

\subsection{Probability Estimation} 
\label{sect:pro}
Suppose the total experimental budget is $ n=40 $. The size of the initial design is kept as $n_0 =20 $,
but two choices are considered for the type of design.
The first option simply generates 20 points at random from the independent input distributions. 
The second choice starts with a random Latin hypercube design \citep[LHD,][]{randomLHS} 
in three variables on $[0, 1]^3$ 
and then maps them uniformly onto the ranges given by the mean $ \pm 3$ standard deviations of the actual distributions.
Hence, the 20 values of the first input, $x_m$, are uniform on $[2000 - 3 \times 400, 2000 + 3 \times 400]$;
similarly the second input.
The values of the lognormal third input just require exponentiating after analogous operations. 
The two types of initial design are called ``random'' and ``uniform''. 

We can see from (\ref{d3_e}) that large $x_m$, large $x_p$, and small $x_z$ tend to lead to failure. 
Hence, we speculate that the uniform initial design will lead to better performance 
in estimation of the failure probability, 
as it over-samples the tails of each distribution where failures occur.
We suspect that over-sampling of extremes and combinations of extremes in the training set  
is desirable for many applications, to identify failure modes. 

The MC set is $100,000$ points independently generated at random from the input distributions. 
The optimization in (\ref{eqn:max:EI}) or (\ref{pro}) is done by additionally generating $ 10,000 $ points from the input distributions to form a finite candidate set and adding the point that maximizes EI or the discrepancy criterion in the candidate set. The whole process is repeated 10 times.  
These sample sizes are meaningfully chosen such that they keep a balance between the estimation accuracy and the computational time. 

\begin{figure}[htbp!]
\centering
\includegraphics[angle=0, width=7in, height=5in]{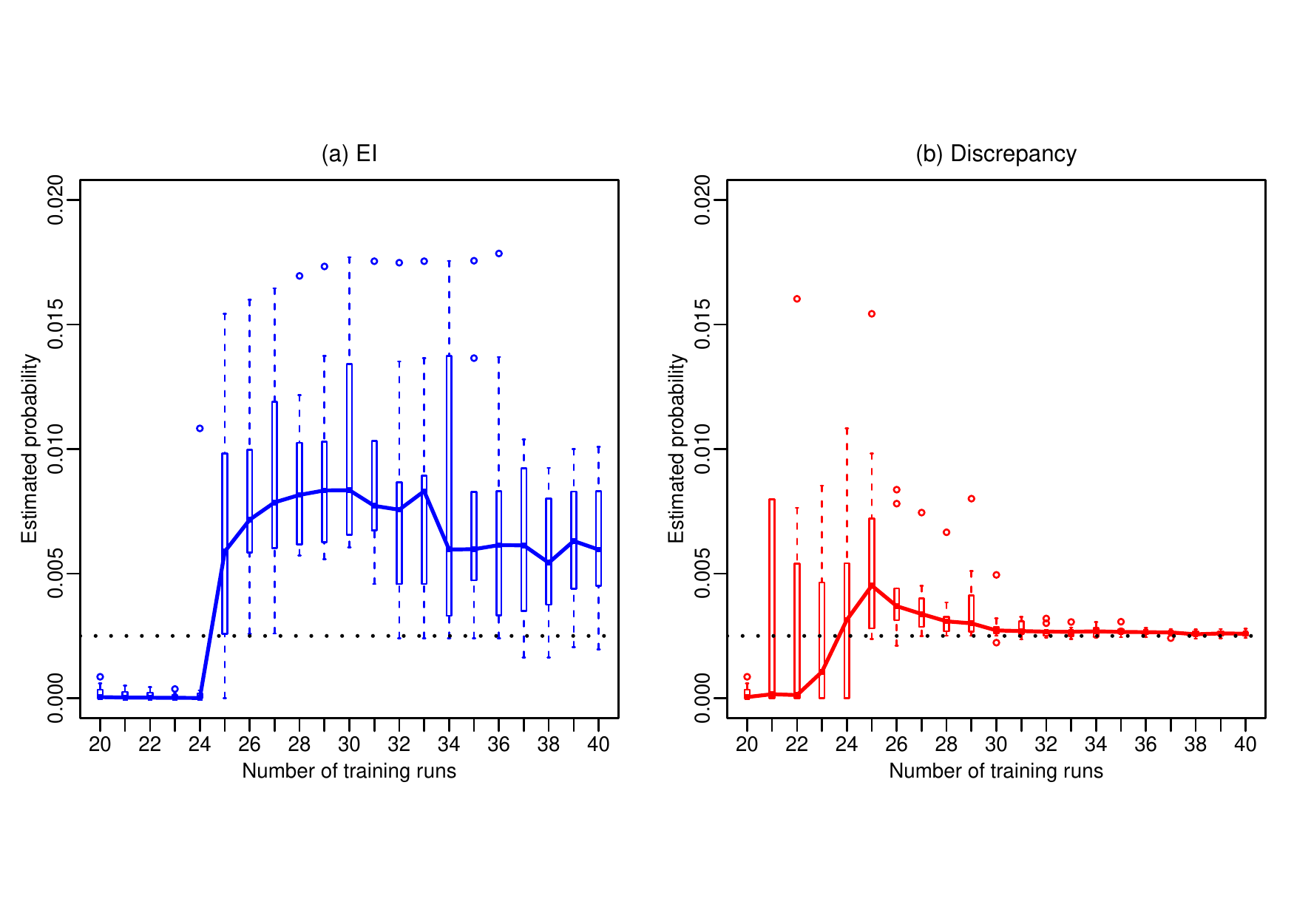} 
\vspace*{-2cm}
\caption{Probability estimates for the short column function when the initial design is random
for two search criteria: (a) EI and (b) Discrepancy. 
The boxplots show $10$ repeat experiments, and the medians are joined by solid lines. 
The dotted line is the true failure probability, $ 0.0025$.}   
\label{d3_10_random}
\end{figure}

Figure \ref{d3_10_random} shows the results starting from random initial designs
(the same 10 initial designs are used for the EI and discrepancy criteria).
Estimates from the EI search barely show any sign of convergence. 
In contrast, after adding a further $11$ points, 
the estimates based on the discrepancy search criterion essentially converge to the true probability. 
They stay there and do not diverge for all 10 repeats of the experiment. 
The root mean squared error (RMSE) over the 10 repeats 
summarized in the first row of Table \ref{d3_10_table} 
confirms that the discrepancy criterion performs 
better than EI for the random initial design.
Further calculations, not reported here for brevity,
show that the EI method  requires about 30 extra points after the 20-point initial design,  
to converge.

\begin{table}[htbp!]
 \centering
 \caption{RMSE of the final ($n = 40$) probability estimate for the short column function.}  
  \begin{tabular}{ccc} \hline
  Initial design & EI    & Discrepancy  \\ \hline
   Random & $ 0.00439 $ & $ 0.00014 $\\ 
    Uniform & $ 0.00094 $   &  $ 0.00011 $ \\ \hline
  \end{tabular}%
  \label{d3_10_table}%
\end{table}%

\begin{figure}[htbp!]
\centering
\includegraphics[width=7in, height=5in, angle=0]{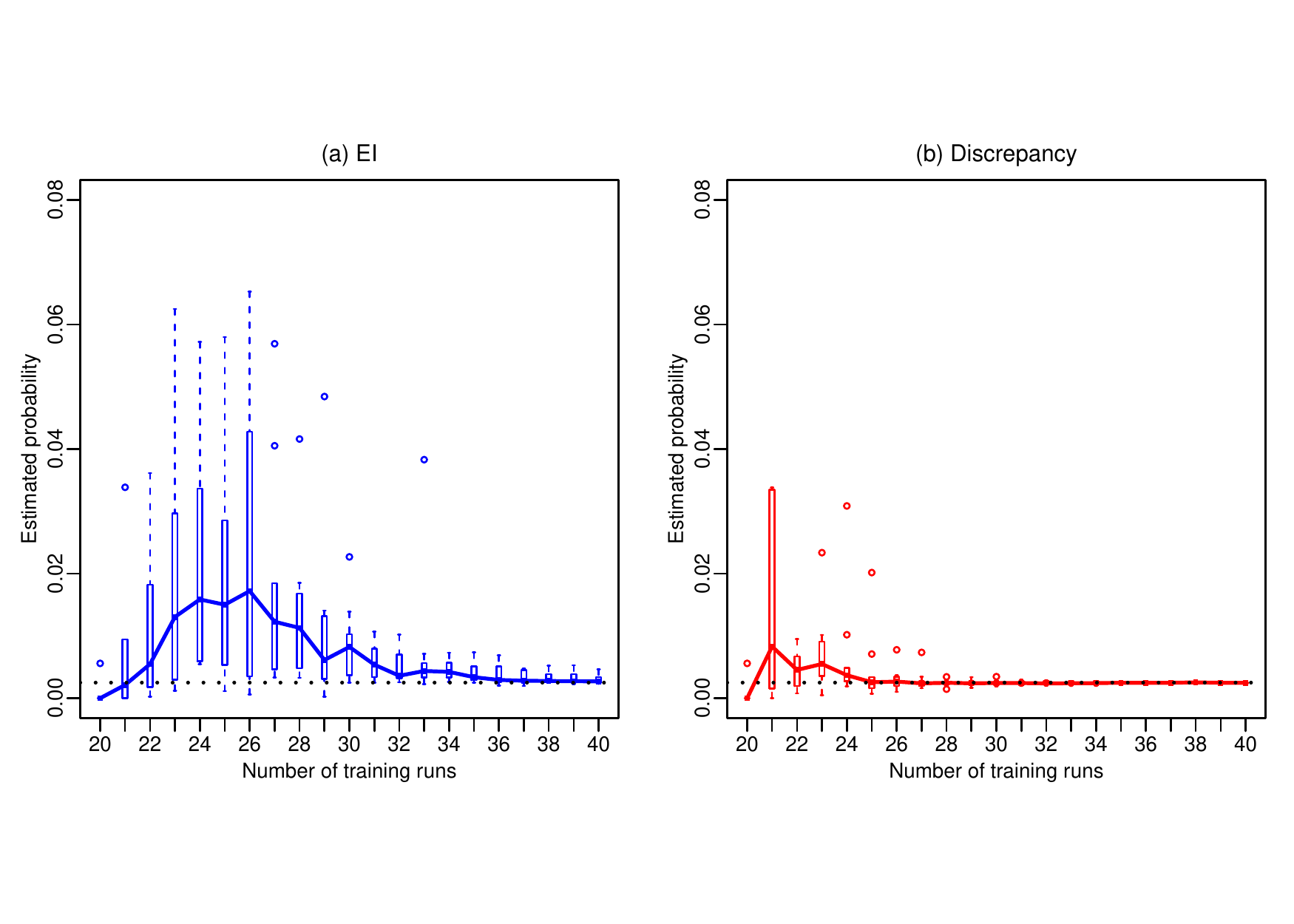} 
\vspace*{-2cm}
\caption{Probability estimates for the short column function when the initial design is uniform
for two search criteria: (a) EI and (b) Discrepancy. 
The boxplots show $10$ repeat experiments, and the medians are joined by solid lines. 
The dotted line is the true failure probability, $ 0.0025$.}   
\label{d3_10_uniform}
\end{figure}

Figure \ref{d3_10_uniform} tells a similar story for uniform initial designs:
the discrepancy criterion outperforms EI. 
The second row of Table \ref{d3_10_table} confirms this. 
In addition, if we compare Figures \ref{d3_10_random} and \ref{d3_10_uniform}, 
the uniform initial designs perform better than the random initial designs, 
in agreement with our conjecture. 
Taking a uniform initial design is especially helpful for the EI criterion.

\begin{figure}[htbp!]
\centering
\includegraphics[width=7in, height=5in, angle=0]{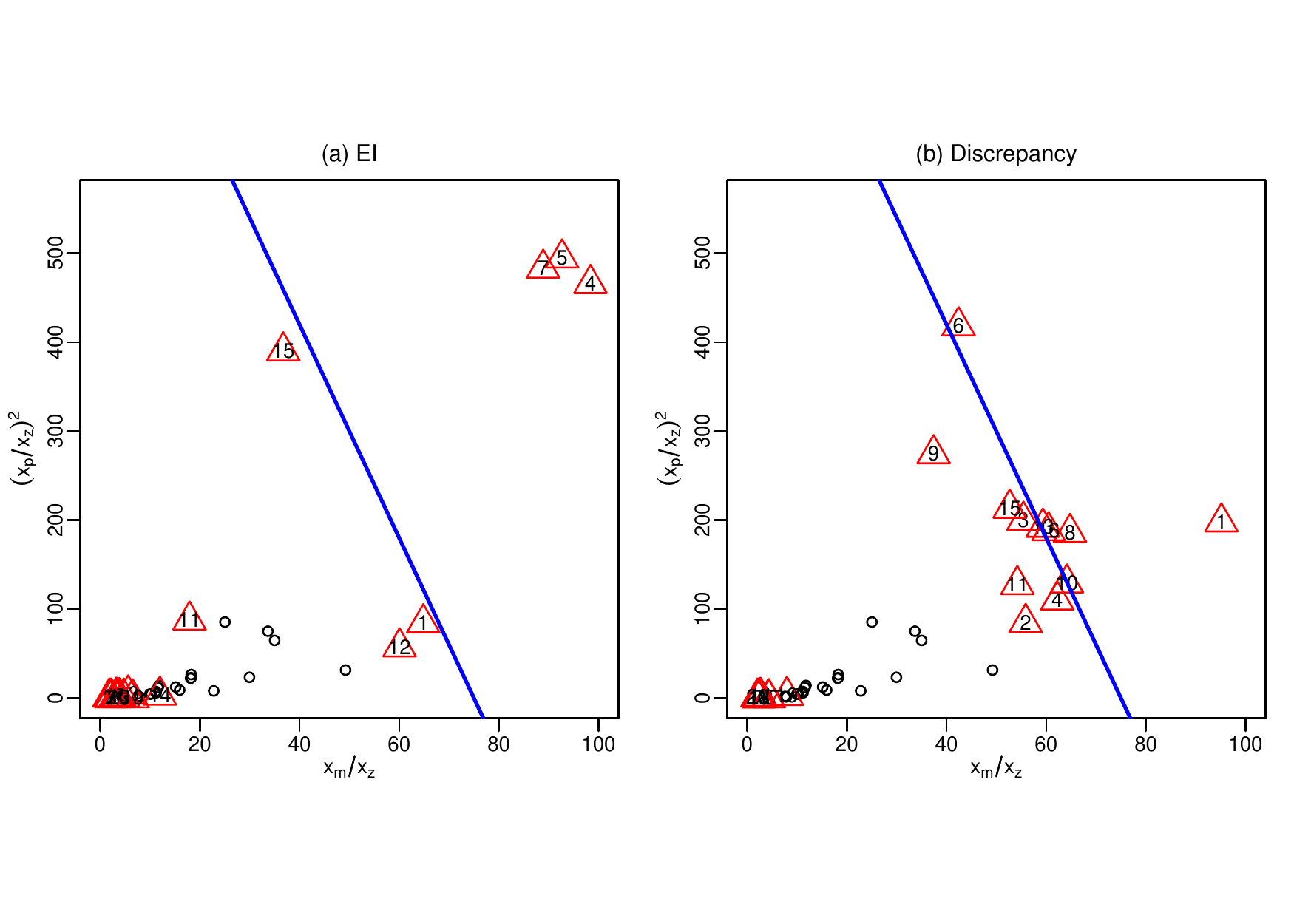}
\vspace*{-2cm}
\caption{Search points for estimation of failure probability for the short column function
as a function of $x_m / x_z$ and $(x_p / x_z)^2$: 
the initial random 20-point design (dots), the failure boundary (solid line), 
and the $20$ points added (triangles with sequence order inside) using (a) EI and (b) discrepancy.}
\label{d3_added}
\end{figure}

The superior convergence using the discrepancy criterion relative to EI 
is next explored by examining the points chosen by the two criteria.
Visualization is facilitated by defining 
$ t_1 = x_m / x_z$ and $t_2=(x_p / x_z)^2$ and noting that the function in (\ref{d3_e}) 
can be rewritten as 
$$ y(\xVec)=1-\frac{4}{b h^2} t_1-\frac{1}{b^2 h^2} t_2. $$
Thus the failure boundary is a function of just two variables, $ t_1 $ and $ t_2 $,
and is linear in them.
For one of the repeat experiments adding 20 points, 
Figure \ref{d3_added} shows the (common) initial points in the $(t_1, t_2)$ subspace, 
the failure boundary of interest, and the $20$ extra points. 
We observe that the initial points are far from the contour of interest, 
but the sequential methods explore the space and, in particular, the contour. 
Among the  $20$ points added, however, the discrepancy criterion places $ 13 $ points 
near the failure boundary, whereas the EI method chooses the majority of the points
in less relevant regions.           

\subsection{Quantile Estimation}
\begin{figure} [htbp!]
\centering 
\includegraphics[width=7in, height=5in, angle=0]{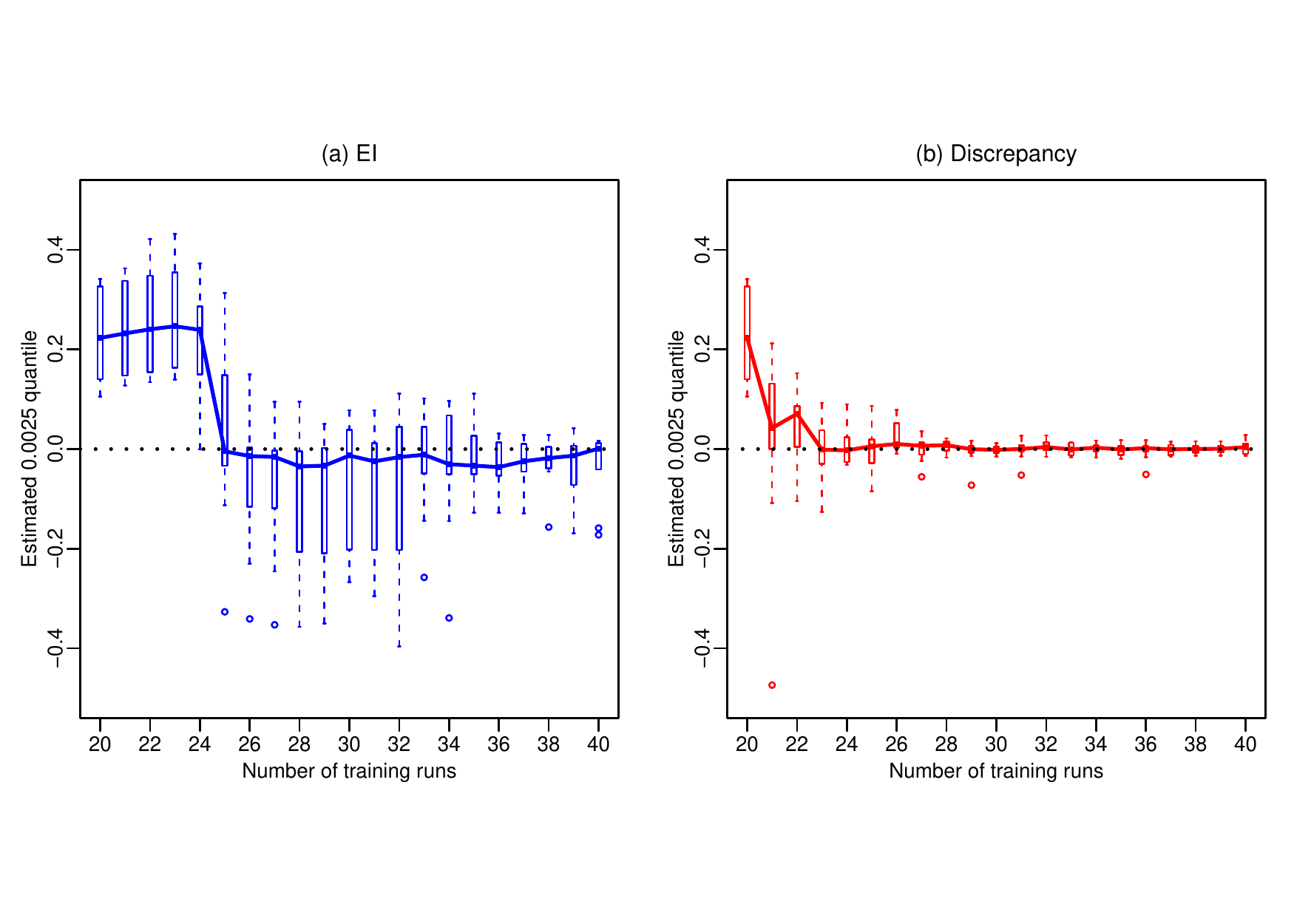} 
\vspace*{-2cm}
\caption{Quantile estimates for the short column function when the initial design is random
for two search criteria: (a) EI and (b) Discrepancy. 
The boxplots show $10$ repeat experiments, and the medians are joined by solid lines. 
The dotted line is the true quantile, 0.}    
\label{d3_quantile_random}
\end{figure}

\begin{figure} [htbp!]
\centering 
\includegraphics[width=7in, height=5in, angle=0]{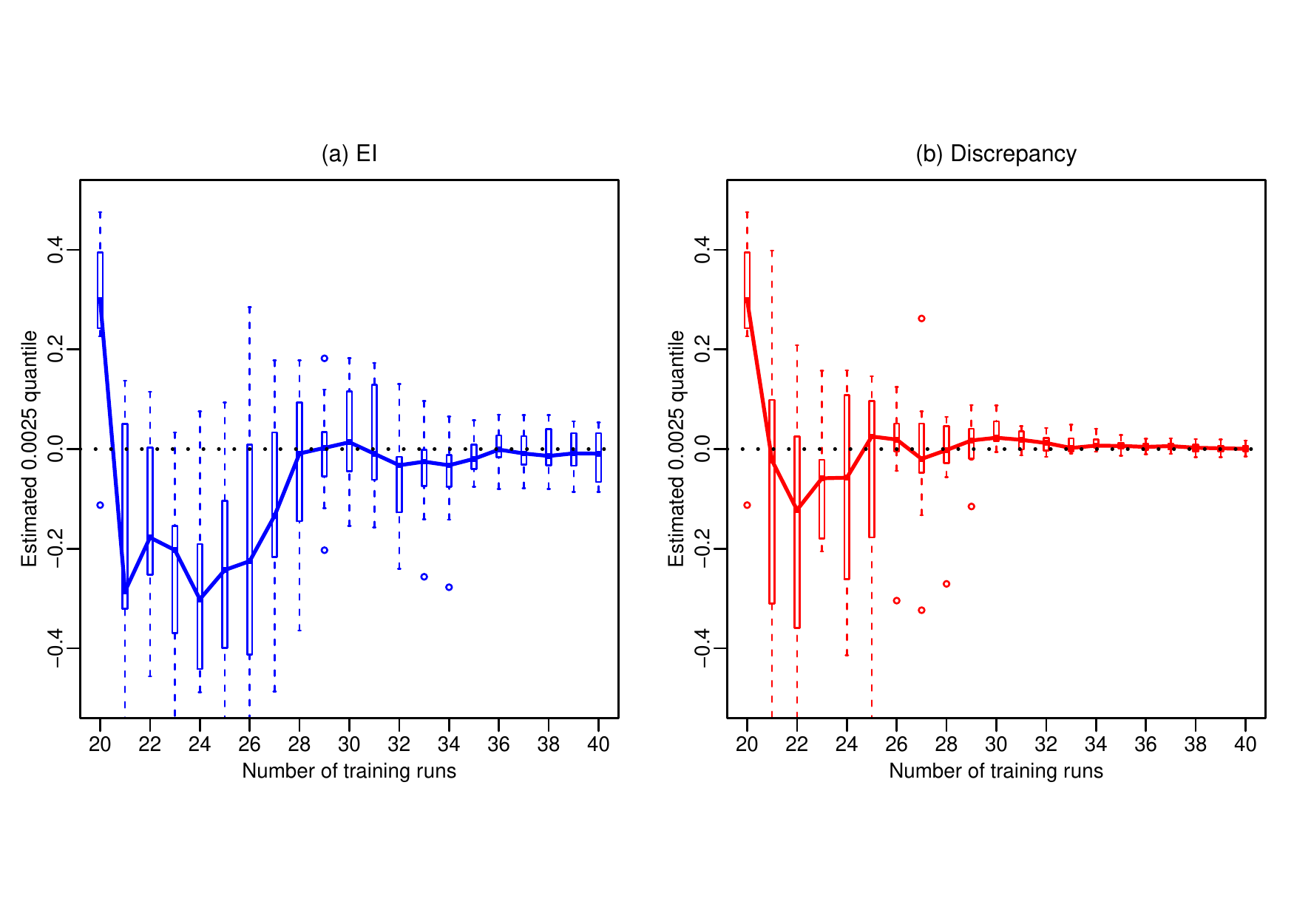}
\vspace*{-2cm}
\caption{Quantile estimates for the short column function when the initial design is uniform
for two search criteria: (a) EI and (b) Discrepancy. 
The boxplots show $10$ repeat experiments, and the medians are joined by solid lines. 
The dotted line is the true quantile, 0.}   
\label{d3_quantile_uniform}
\end{figure}

Following the same settings as in Section \ref{sect:pro}, we investigate extreme quantile estimation. 
Results are reported in  Figure \ref{d3_quantile_random} for random initial designs, 
where we see the discrepancy criterion again providing much faster convergence, 
similarly in Figure \ref{d3_quantile_uniform} for initial uniform designs.
While comparison of Figures~\ref{d3_quantile_random} and \ref{d3_quantile_uniform} shows
that the initial uniform designs perform poorly at the outset and when just a few points are added sequentially, the RMSE values in Table \ref{d3_10_table_quantile} indicate that the uniform designs ultimately provide a better solution for both search criteria at $n = 40$.  
      
\begin{table}[htbp!]
 \centering
 \caption{RMSE of the final ($n = 40$) quantile estimate for the short column function.}  
  \begin{tabular}{ccc} \hline
  Initial design & EI    & Discrepancy  \\ \hline
   Random & $ 0.07592 $ & $ 0.01287 $\\ 
    Uniform & $ 0.04999 $   &  $ 0.00991 $ \\ \hline
  \end{tabular}%
  \label{d3_10_table_quantile}%
\end{table}%

Thus, the results for probability estimation and quantile estimation lead to the same conclusions. 
The discrepancy criterion gives estimates that converge faster than those for the EI method for both initial designs. A uniform initial design tends to outperform a random initial design. Based on these observations, we recommend the discrepancy criterion starting from a uniform initial design.

\section{Application to a Computer Model of a Floor System}
\label{sect:wood}

\subsection{Preliminary Analysis}
\label{sect:pre}
Consider the computer model introduced in Section \ref{sect:floor:model} with $ d=8 $. Before doing any formal modelling, we carry out a preliminary sensitivity analysis by fitting a constant mean Gaussian process with training data from an initial random LHD with $ n=20 $. We use functional analysis of variance (ANOVA) which takes each variable's main effect as well as all of the two level interaction effects into account \citep{SchWel2006}. All higher order interactions are ignored. The ANOVA decomposition results are reported in Table \ref{tab:ANOVA}.

\begin{table}[htbp!]
\caption{Functional ANOVA for the floor-system application showing the percentage contributions
from the eight main effects to the total variation of the GP prediction over the $8$-dimensional input space. The main effects explain $99.2 \%$ of the total variance in the GP prediction. 
Interactions between two input factors are not shown as none of them contributes more than 0.5\%.} 
\begin{center}
\begin{tabular}{ |c|c| } 
\hline
Input (MOE) & \% variation \\  
 \hline
$x_1$ & 0.0  \\ 
$x_2$ & 5.3  \\ 
$x_3$ & 12.8  \\ 
$x_4$ & 26.2  \\ 
$x_5$ & 37.5  \\ 
$x_6$ & 16.1  \\ 
$x_7$ & 1.0   \\ 
$x_8$ & 0.2  \\ 
 \hline
\end{tabular}
\label{tab:ANOVA}
\end{center}
\end{table}

\begin{figure}[htbp!]
\centering
\includegraphics[height=5in, angle=0]{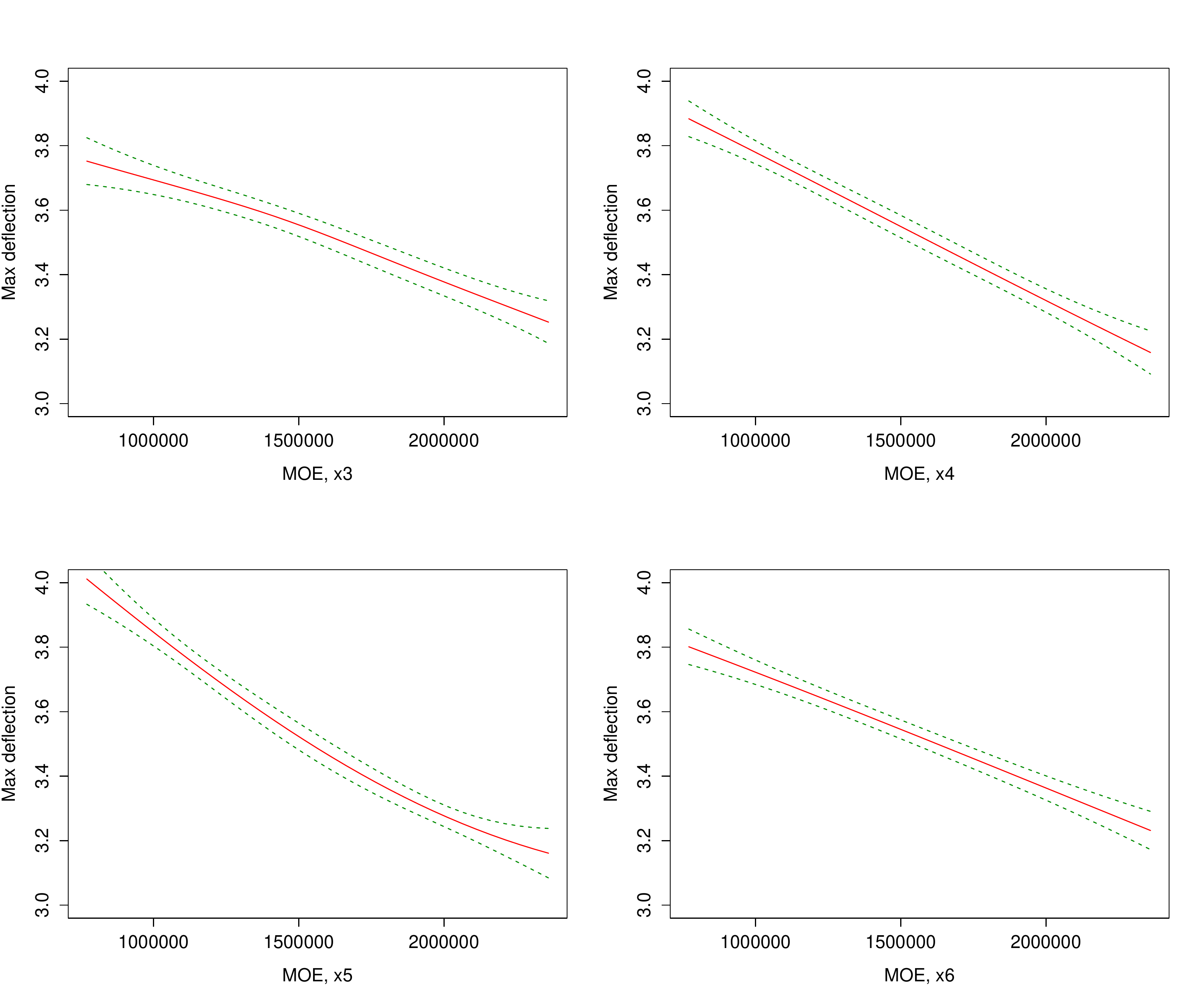} 
\caption{The relationship between the MOEs of the four middle beams and the response. 
The solid lines are the estimated main effects, with 95\% pointwise confidence intervals as dotted lines. Note that the inputs are between $ 0.77 \times 10^6 $ and $ 2.36 \times 10^6 $ (pounds per square inch), which will be explained in Section \ref{sect:input}.}
\label{influence}
\end{figure}

From Table \ref{tab:ANOVA}, we observe that the four middle beams have the most important effects on the predicted response, $ y $. The relationship between them and the response is illustrated in Figure \ref{influence}, again using the methods of \cite{SchWel2006}. The relationship between the four side beams and $ y $ are similar to those of the four middle beams but weaker. It is clear from Figure \ref{influence} that a small MOE value results in a larger deflection. Therefore, the lower tail of the input distribution is important when estimating the upper probability. This is valuable prior information obtained from the preliminary analysis.   

In the short column function of Section \ref{sect:toy}, the function is explicitly specified in (\ref{d3_e}), and we can determine immediately the critical part of the input space. In practice, however, when the functional relationship between inputs and output is unknown, we can carry out a preliminary study as in this section to examine the effects of the inputs.

\subsection{Modelling the Input Distribution} 
\label{sect:input}
\begin{figure}[htbp!]
\centering
\includegraphics[width=5in, height=5in, angle=0]{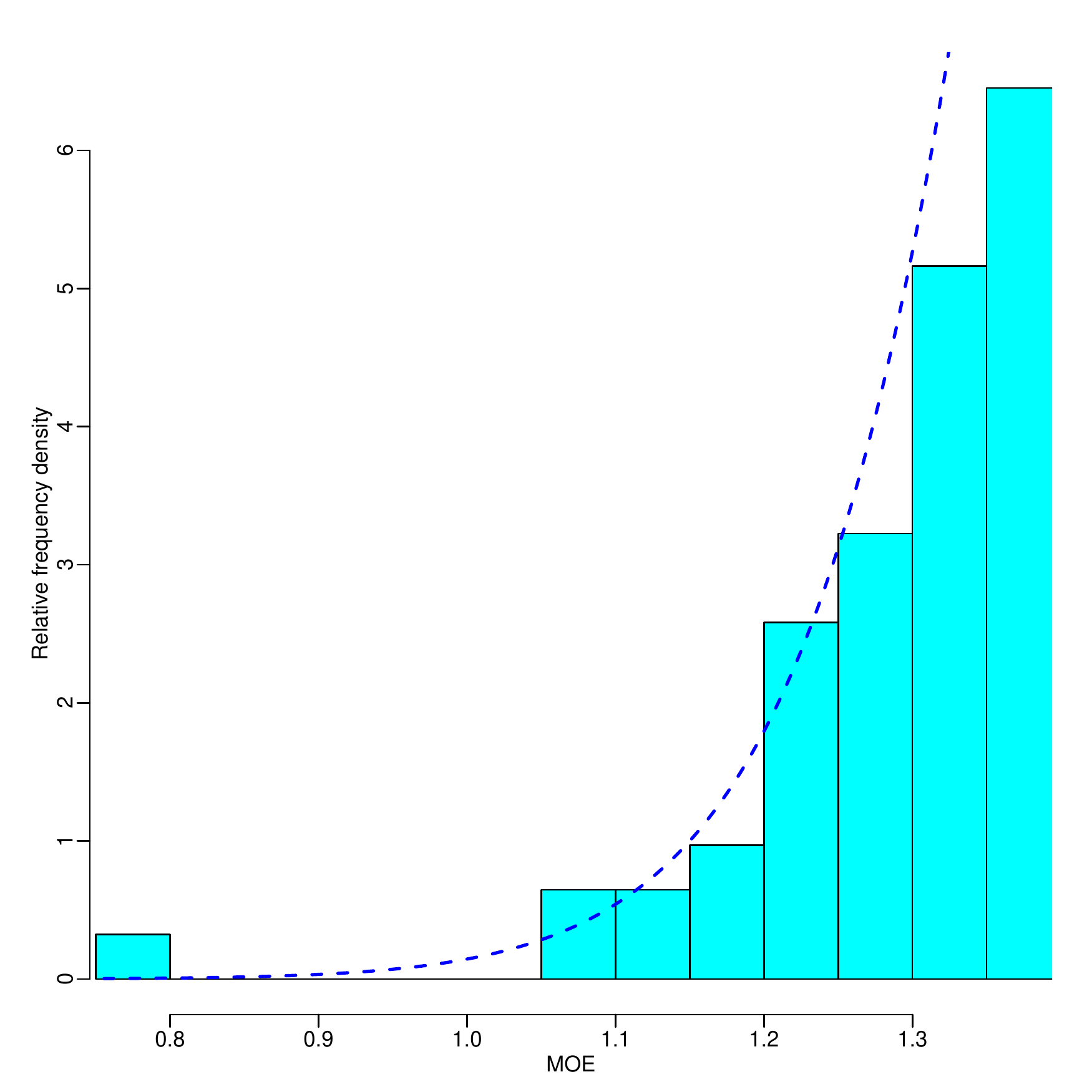}
\caption{Modelling the lower 10\% of the MOE data: Histogram of the lower 10\% of the data and
a fitted 2-parameter (or 3-parameter) censored Weibull distribution.}
\label{censored}
\end{figure}

We need to specify the input distribution. From the analysis in Section \ref{sect:pre}, we know that a small MOE leads to a large deflection. Therefore, the lower tail of each marginal input distribution is critical. A dataset of $580$ MOE values measured from $580$ boards collected from production is available to us. One way to proceed is to fit a parametric input distribution, for example the Weibull. However, training a parametric distribution based on all of the $ 580 $ MOE data values does not emphasize the importance of the lower tail. Therefore, initially we considered the following two semi-parametric input distributions: 
\begin{itemize}
\item a mixture distribution, with probability $p_1=0.1$ to sample from a 2-parameter censored Weibull distribution and $ p_2=0.9 $ to sample with replacement from the upper 90\% of the empirical distribution of the available data. The 2-parameter censored Weibull distribution is trained using complete data from the lower 10\% of the available data with the rest (90\% of the data) treated as right censored \citep{liu2018lower}. 

\item As above but using a 3-parameter Weibull distribution for the first stratum. 
\end{itemize}
The unknown parameters were estimated by MLE.
The resulting estimated density curves for the 2- and 3-parameter Weibull distributions
are virtually identical and added to the empirical histogram of the lower 10\% data shown in Figure \ref{censored}.
It is clear that there is a non-negligible bump (3 boards) in the lower tail of the dataset, which neither of the semi-parametric distributions is able to characterize. 

Hence, we propose to use instead the following non-parametric input distribution, $ H(x) $: 
\begin{equation} \label{mixturedis}
H(x) =  p_1\times G_1(x) + p_2 \times G_2(x), 
\end{equation}
where $G_1(x) $ is the empirical distribution of the lower 10\% of the MOE data,
and $G_2(x)$ is the empirical distribution of the upper 90\% of the data. 
The weights $p_1 = p_2 = 0.5$ deliberately over-sample the lower tail of the input distribution. They will later be corrected using methods for weighted stratified random sampling when computing the probability/quantile estimates. In a nutshell, we use the empirical distribution of the 580 MOE data values as the input distribution, but deliberately over-sample the lower tail when generating the MC set and the finite candidate set.  
 
\subsection{MC Set} \label{mcset}
The MC set is comprised of $ 12,800 $ points generated as follows. Each dimension has two strata for the mixture distribution: either sampling from $G_1(x)$ or sampling from $G_2(x) $. 
Hence, in total there are $ 2^8=256 $ strata. For each of the $256$ strata, we generate $ 50 $ different points from the relevant combination of $ G_1(x) $ and $ G_2(x) $ distributions and thus we have $  256 \times 50 = 12,800 $ points in total to form the MC set. The above sample sizes are meaningfully chosen such that they keep a balance between the approximation accuracy and the computational time.  

Working with this MC set, step 6 in Algorithm \ref{alg:p} for probability estimation will change to 
\begin{equation} \label{prob_MC_set}
\hat{p}_{f} = \sum_{h=1}^{256} w_h \frac{1}{50} \sum_{i = 1}^{50} \mathds{1} (\hat{y}(\xVecMC^{(hi)}) > y_f) ,
\end{equation}
where $\xVecMC^{(hi)})$ is MC sample point $i$ in stratum $h$, and the stratum weights $ w_h $ sum to 1. 
For instance, the combination with all inputs coming from $G_1(x)$ has $ w_h = (0.1)^{8}$ as the stratum weight. For quantile estimation, step 6 in Algorithm \ref{alg:q} also changes to finding $\hat{y}_{f} $ such that $  \hat{p}_{f} $ equals a pre-specified probability, $ a $. In practice, any trial value of $y_f $ gives a $ \hat{p}_f $ in (\ref{prob_MC_set}). We minimize $ |  \hat{p}_{f} - a| $ numerically with respect to $y_f $ using the optimize() function in the MASS package of R to find $ \hat{y}_{f} $ with $10^{-6} $ tolerance. 

\subsection{True Probability and Quantile}
With $ p_{f} = 0.001$, we simulated 10 different MC sets from the mixture distribution in (\ref{mixturedis}) and ran the computer model. With 10 repeats the mean of the estimates of the $ 0.999 $ quantile is $ 3.88057 $ inches and the standard error of the mean is $0.00701$. Therefore, the ``true'' $0.999$ quantile is taken to be 3.88 inches.

\subsection{Application Results} 
\label{sect:results}
Suppose the design budget is $ n=30 $. The number of points for the initial design is $ n_0=20 $ and $10$ additional points are chosen sequentially by optimizing a search criterion. The whole process is repeated $ 10 $ times with different initial designs. The MC set contains 12,800 points simulated as described in Section \ref{mcset}. The finite candidate set is the same as the MC set. For the initial design with only $n_0=20$ points, however, we take a random uniform LHD to cover the input space globally, following the recommendations from the short column model in Section \ref{sect:toy}. 

\begin{figure}[htbp!]
\centering
\includegraphics[width=7in, height=5in, angle=0]{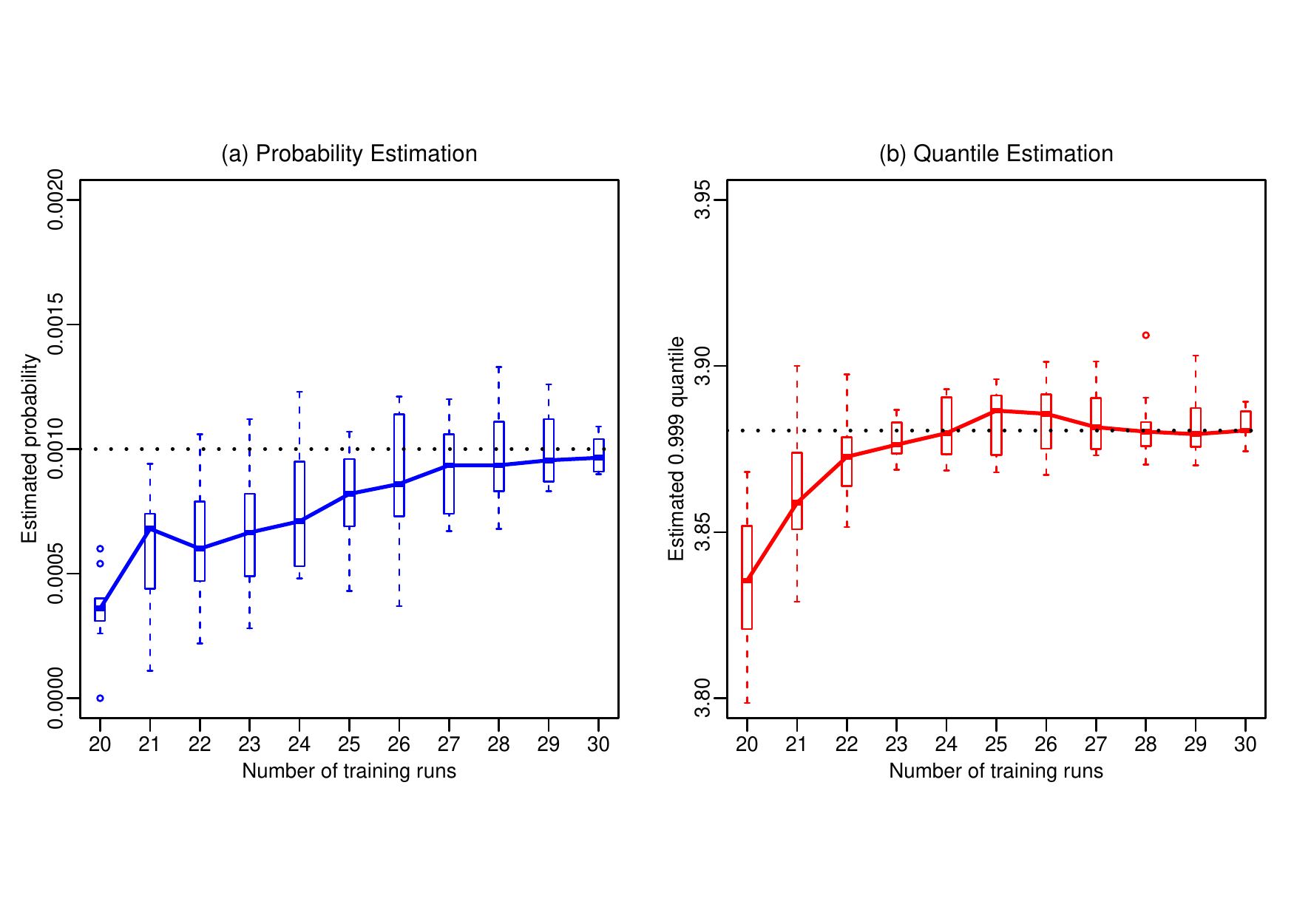} 
\vspace*{-2cm}
\caption{Computer model of the floor system with a uniform initial design of $n = 20$ points
and sequential search using the discrepancy criterion: (a) probability estimation and (b) quantile estimation. The boxplots show $10$ repeat experiments, and the medians are joined by solid lines. The dotted lines are the true probability $ 0.001$ and the true quantile $3.88$, respectively.}
\label{wood_uniform}
\end{figure}

Results based on the discrepancy criterion are reported in Figure \ref{wood_uniform}. 
The initial estimates of the tail probability are typically only about 30\% of the true value of 0.001
but vary tightly around the truth after 10 additional points, The quantile estimates similarly converge to the true quantile. Moreover, the extreme probability and quantile are estimated well using just the carefully designed 12,800 points in the MC and candidate sets and adjusted stratum weights. This computational efficiency is a result of the preliminary analysis that established a negative association between the MOEs of the joists and the associated deflection.


\section{Diagnostics} \label{set:diagg}
In practice, the true tail probability or quantile is unknown, 
yet a user still wants guidance about whether the algorithm has converged.
Furthermore, the question of convergence relates to the specific single experiment
actually executed.
Here we describe a diagnostic for convergence of a sequential search based on the discrepancy criterion.

From the formulation in (\ref{pro}), the negative of the absolute discrepancy is 
$-|\hat{m}_{\psiVec}( \xVec)-y_f| / \sqrt{v_{\psiVec,\sigma^2}(\xVec)}$.
Ideally, it  will become more negative for all points in the MC set as the training set increases. 
Thus, we compute this criterion point-wise for the MC set at each step 
and draw a boxplot against the sample size. 
We would expect a declining trend with sample size.
 
\begin{figure} [htbp!]
\centering 
\includegraphics[width=7in, height=5in, angle=0]{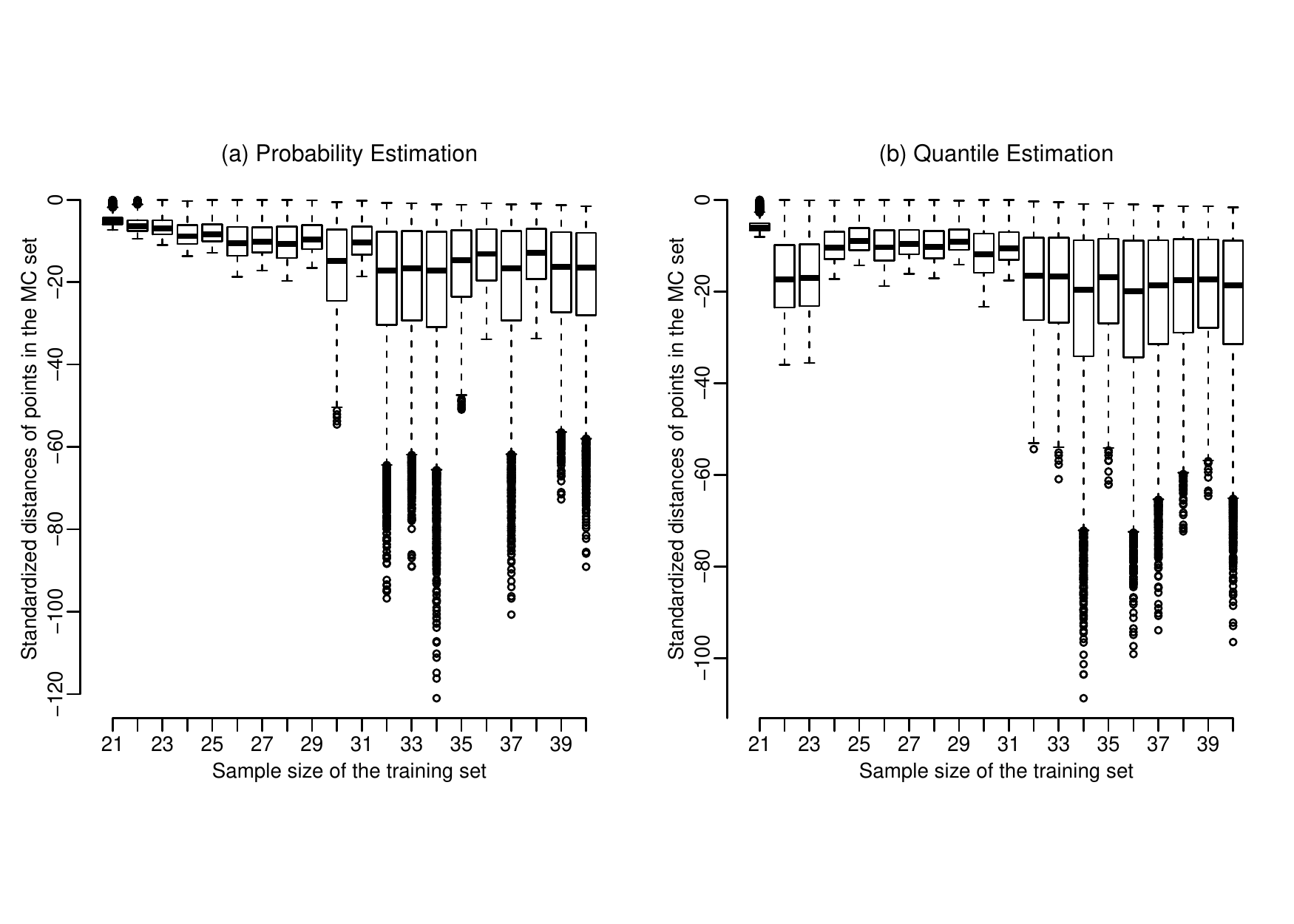}  
\vspace*{-2cm}
\caption{Diagnostic plots for the short column function starting with a uniform initial design: 
(a) probability estimation and (b) quantile estimation.} 
\label{d3_diag_short}
\end{figure}

\begin{figure} [htbp!]
\centering 
\includegraphics[width=7in, height=5in, angle=0]{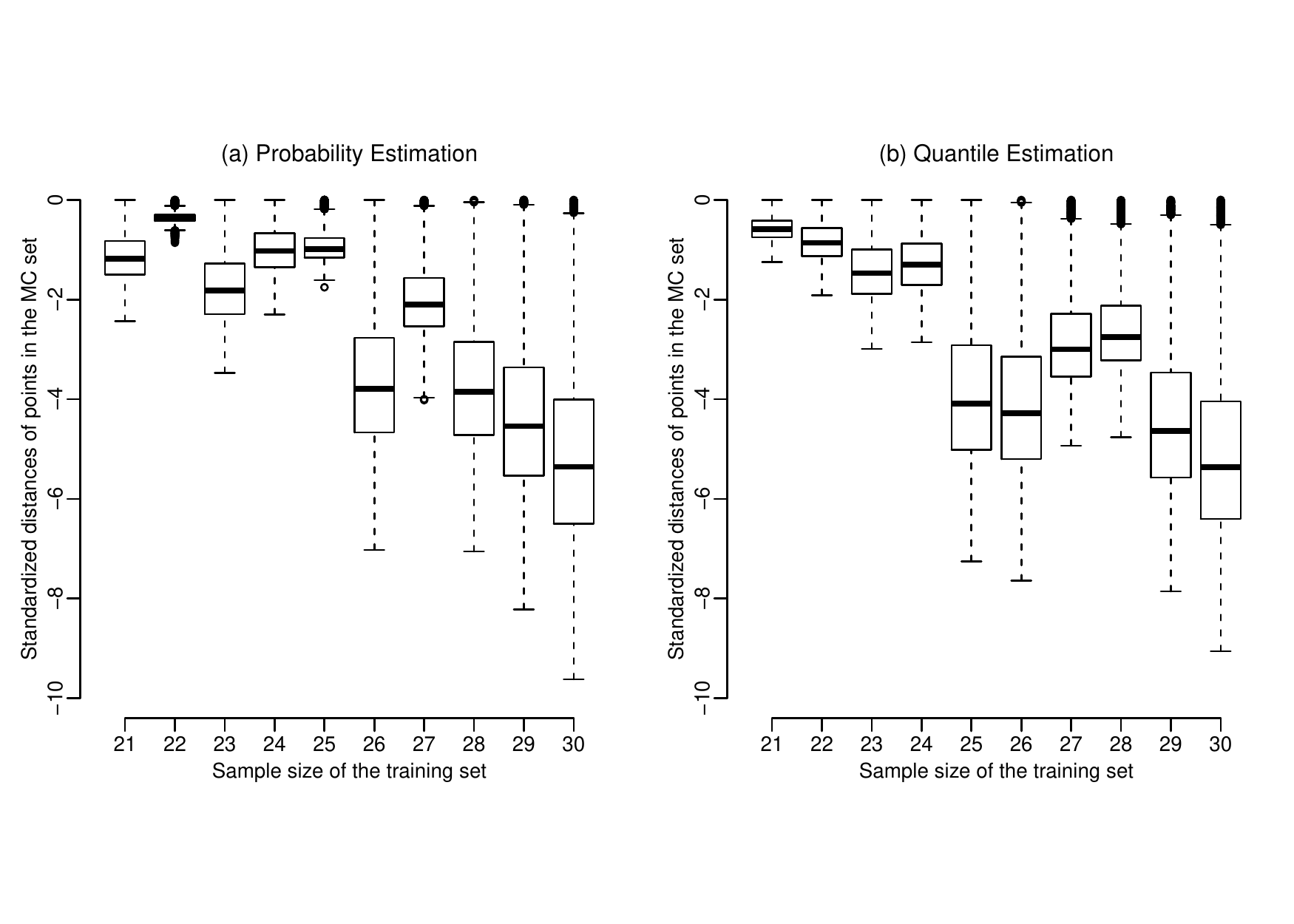}  
\vspace*{-2cm}
\caption{Diagnostic plots for the floor system computer model starting with a uniform initial design: 
(a) probability estimation and (b) quantile estimation.} 
\label{d3_diag_wood}
\end{figure}

Figures \ref{d3_diag_short} and \ref{d3_diag_wood} show  diagnostic plots for one repeat of
the experiments with 
the short column function and the floor system computer model, respectively.
Both experiments start from a uniform initial design.

Consistent with our expectation, both plots exhibit a decreasing trend. 
After adding 10 points to the initial design for the short column function, 
most of the discrepancies in the MC set are less than $-10$, 
i.e., a huge standardized distance between the predictive mean and the contour of interest. 
The pattern carries over to the computer model for the floor system, 
which again indicates the algorithm is converging.

\section{Concluding Remarks}   \label{sect:remarks} 
This paper illustrates the usefulness of a sequential strategy to estimate an extreme probability or its associated quantile. Throughout the paper, we have addressed some practical issues an engineer faces when using a sequential design. From  the analyses, we have the following recommendations. 
\begin{itemize}
\item Use the discrepancy criterion, which is more straightforward and converges faster than EI in our study. 
\item Use a uniform initial design, to over-sample the tails of the input distributions 
(for the floor model only one tail is over-sampled).  
\item  It may be more efficient to choose the MC set and the candidate set according to a stratified weighting scheme,
to enrich the sets with points in the failure region. 
That is more efficient than generating huge sets from the input distributions. 
\item It can be important to do a preliminary analysis to gain some insights on the relationship between inputs and output. 
\item The search criteria considered in this paper use the standard error of prediction to guide local/global search.  As Bayesian methods can provide more realistic uncertainty estimates \citep{chen2017flexible}, their use is recommended for sequential search. 
\end{itemize}

\section*{About the authors}

\noindent
\censor{Hao Chen was a PhD candidate in the Department of Statistics at the time of this research. 
He is currently a senior research scientist at R\&D, LoyaltyOne Inc.}

\noindent
\censor{William J.\ Welch is a Professor in the Department of Statistics, University of British Columbia.}

\section*{Funding}

\noindent
\censor{Research supported by the Natural Sciences and Engineering Research Council, Canada
and Compute Canada/WestGrid.}

\bibliographystyle{asa}
\bibliography{sample}

\end{document}